%% file: main.tex
\documentclass{article}

\usepackage{arxiv}
\usepackage[utf8]{inputenc} 
\usepackage[T1]{fontenc}    
\usepackage{hyperref}       
\usepackage{url}            
\usepackage{booktabs}       
\usepackage{amsfonts}       
\usepackage{nicefrac}       
\usepackage{microtype}      
\usepackage{graphicx}
\usepackage{natbib}
\usepackage{doi}

\usepackage{amsmath}
\usepackage{amsthm}
\usepackage{algorithm}
\usepackage{algorithmic}
\usepackage[switch]{lineno}

\newtheorem{theorem}{Theorem}

\usepackage{microtype}
\usepackage{subfigure}
\usepackage{amsfonts}
\usepackage{multirow}
\usepackage{mathtools}
\usepackage{makecell}
\usepackage{latexsym}
\usepackage{amssymb}
\usepackage{multicol}
\usepackage{tablefootnote}
\usepackage{arydshln}
\usepackage{wrapfig}
\usepackage{multirow}
\usepackage[normalem]{ulem}
\useunder{\uline}{\ul}{}
\newcommand{\eref}[1]{Eq.~(\ref{#1})}

\newcommand{\cref}[1]{Condition~(\ref{#1})}

\newcommand{\secref}[1]{Section~\ref{#1}} 
\newcommand{\figref}[1]{Fig.~\ref{#1}} 
\newcommand{\tabref}[1]{Table~\ref{#1}} 

\let\leq\leqslant

\DeclareMathOperator*{\argmin}{arg\,min}
\DeclareMathOperator{\sign}{sign}
\newcommand{\NAME}[0]{SLTR}

\newtheorem{corollary}{Corollary}[section]

\title{Sparse and Low-Rank High-Order Tensor Regression via Parallel Proximal Method}

\date{} 					

\author{
    {\hspace{1mm}Jiaqi Zhang} \\
	Department of Computer Science\\
	Brown University\\
    \And
    {\hspace{1mm}Yinghao Cai} \\
	Department of Computer Science\\
	Southeast University\\
    \And
    {\hspace{1mm}Zhaoyang Wang} \\
	Department of Computer Science\\
	Southeast University\\
    \And
    {\hspace{1mm}Beilun Wang}\thanks{To whom correspondence should be addressed.} \\
	Department of Computer Science\\
	Southeast University\\
}

\begin{document}

\maketitle

\begin{abstract}
Recently, tensor data (or multidimensional array) have been generated in many modern applications, such as functional magnetic resonance imaging (fMRI) in neuroscience and videos in video analysis. Many efforts are made in recent years to predict the relationship between tensor features and univariate responses. However, previously proposed methods either lose structural information within tensor data or have prohibitively expensive time costs, especially for large-scale data with high-order structures. To address such problems, we propose the Sparse and Low-rank Tensor Regression (SLTR) model. Our model enforces sparsity and low-rankness of the tensor coefficient by directly applying  $\ell_1$ norm and tensor nuclear norm, such that it preserves structural information of the tensor. To make the solving procedure scalable and efficient, SLTR makes use of the proximal gradient method, which can be easily implemented parallelly. We evaluate SLTR on several simulated datasets and one video action recognition dataset. Experiment results show that, compared with previous models, SLTR can obtain a better solution with much fewer time costs. Moreover, our model's predictions exhibit meaningful interpretations on the video dataset.
\end{abstract}

\input{sec/intro.tex}
\input{sec/background}

\input{sec/method}
\input{sec/theory}
\input{sec/related_work}
\input{sec/result}
\input{sec/discussion}

\bibliography{citations}
\bibliographystyle{apalike}


\end{document}

%% file: sec/intro.tex
\section{Introduction}
Tensor data, also called multidimensional array data, is frequently seen in various scientific and real-world applications, including neuroscience \citep{zhu2014fusing,noroozi2020tensor}, video analysis \citep{wu2010tensor,lui2012least,yang2017tensor}, and recommendation system design \citep{sharma2013survey,bhargava2015and}. For example, functional magnetic resonance imaging (fMRI) data in neuroscience contains a series of 3D tensors (3-mode or 3-order data) with the shape of \emph{time} $\times$ \emph{neuron} $\times$ \emph{neuron}. The researchers use such data to predict scalar-valued disease symptoms such as Mild Cognitive Impairment \citep{zhu2014fusing}. Many studies analyze the relationship between a tensor variable and its corresponding scalar response through the so-called \emph{tensor regression} \citep{ji2019survey}.

The emergence of tensor data enables researchers to analyze the system with the presence of structural correlations. However, the characteristics of tensor data also present new challenges for statistical analysis. First, the tensor data are usually high-dimensional, so the number of observations is much less than the number of variables. For example, each sample of the CMU2008 fMRI dataset \citep{mitchell2008predicting} is a $51 \times 61 \times 23$ 3D tensor with 71553 voxels. But, only 360 samples are recorded. Conventional methods are mostly likely to fail in high-dimensional settings since we are trying to infer a large number of unknowns with limited observations. Second, tensor data have a high-order structure. For example, in the recommendation system, we need to predict recommendation levels from 4D user activities \emph{user} $\times$ \emph{product} $\times$ \emph{location} $\times$ \emph{timestamp} \citep{bhargava2015and}. Hence, existing linear regression models cannot be directly applied to tensors. Because such a method is designed for vector data and may lose the spatial structure of the data, like pixel relations in pictures or time orders in videos. Moreover, as the number of variables increases exponentially with the number of dimensions, higher-order data contains more unknown variables and requires significantly expensive computational costs.

To address these challenges, recent tensor regression methods adopt sparse and low-rank constraints from vector or matrix regression methods. The sparse constraint filter out "useless'' variables, generally obtained through variable selection \citep{heinze2018variable}, to decrease the number of variables. On the other hand, enforcing low-rank constraints reduces the complexity of the model that fits the data. These constraints make the tensor regression problem more tractable. For example, \citep{zhou2013tensor,he2018boosted} use CANDECOMP/PARAFAC (CP) decomposition to characterize an $M-$order tensor with multiple components and add structural constraints on each. However, all the CP decomposition-based methods suffer the drawbacks of slow convergence \citep{li2013some} and inaccurate prediction since the best CP approximation might not exist \citep{cichocki2016tensor}.

Therefore, some methods directly apply structural constraints on tensors to avoid the decomposition \citep{song2017multilinear,li2019sturm} or use the more flexible Tuck decomposition \citep{ahmed2020tensor}. However, they are computationally expensive because of expensive procedures, including optimizing multiple nuclear norms. Due to their drawbacks, the above methods can not obtain a solution efficiently for large-scale tensor data. Therefore, we need a fast and scalable tensor regression estimator.

In this paper, we propose the \underline{S}parse and \underline{L}ow-Rank High-Order \underline{T}ensor \underline{R}egression (\NAME{}) method. Our model directly applies sparse and low-rank constraints through $\ell_1$ norm and nuclear norm to decrease the model complexity of tensor regression. To speed up the optimization, we also propose a scalable solution, making use of the parallel proximal method \citep{combettes2011proximal} that can be implemented parallelly. Therefore, through multi-threading computation or Graphics Processing Units (GPUs), the optimization of \NAME{} vastly reduces the computational time cost. We theoretically prove the sharp error bound of our model. Moreover, we compare our model with four state-of-the-art tensor regression methods on several simulated datasets and one video action recognition dataset \citep{soomro2012ucf101}. Results show that our \NAME{} can obtain better solutions with much fewer time costs.

\section{Notation}\label{sec:notation}
We let calligraphic characters denote an $M$-order tensor $\mathcal{A} \in \mathbb{R}^{p_1 \times \cdots \times p_M}$ with the size of each dimension as $\mathcal{P}=\{p_1, \cdots, p_M\}$.  Uppercase characters $A$ denote matrices and lowercase characters $a$ denote vectors. $||\cdot||_1$ and $||\cdot||_\infty$ represent element-wise $\ell_1$ norm element-wise $\ell_\infty$ norm correspondingly. Moreover, $||\cdot||_F$ is the element-wise $\ell_2$ norm (Frobenius norm). For a matrix, $||\cdot||_2$ and $||A||_*$ denote the spectral norm and nuclear norm. We also use the same notations for nuclear and spectral norms of tensors. They can be distinguished based on context. 

We introduce some basic operations for the tensor data. The inner product for two tensor $\mathcal{A}, \mathcal{B} \in \mathbb{R}^{p_1 \times \cdots \times p_M}$ is the sum of products of every entries, defined as $\langle \mathcal{A}, \mathcal{B} \rangle = \sum_{i_1=1}^{p_1}\cdots\sum_{i_M=1}^{p_M}\mathcal{A}_{i_1\cdots i_M}\mathcal{B}_{i_1\cdots i_M}$. The $m-$mode product of an $M-$order tensor $\mathcal{A}$ by a matrix $A \in \mathbb{R}^{J \times p_m}$, denoted by $\mathcal{A} \times_m A$, is a tensor with the shape $\mathbb{R}^{p_1 \times \cdots p_{m-1} \times J \times p_{m+1} \times \cdots \times p_M}$. Here, each entry of the $m-$order product is given by $(\mathcal{A} \times_m A)_{i_1 \cdots i_{m-1} j i_{m+1} \cdots i_M} = \sum_{i_m=1}^{p_m} \mathcal{A}_{i_1 \cdots i_M}a_{ji_m}$. 

%% file: sec/background.tex
\section{Background and Formal Problem Statement}\label{sec:background}

\subsection{Tensor Regression}

In this paper, we consider modelling the relationship between $M$-order tensor variable $\mathcal{X}_i \in \mathbb{R}^{p_1 \times \cdots \times p_M}$ and corresponding scalar response $y_i$ from $N$ observations ($i=1,2,\cdots, N$). We assume a linear relationship as
\begin{equation}\label{eq:tensor-linear-assumption}
    y_i = \langle \mathcal{W}, \mathcal{X}_i \rangle ~+~ \gamma_i,~\forall i=1,\cdots, N,
\end{equation}
where $<\cdot,\cdot>$ is the tensor inner product operator and $\gamma_i \in \mathbb{R}$ is the noise assumed to be drawn from a Normal distribution $\mathcal{N}(0, \alpha)$ with a relatively small $\alpha$. The $M$-order coefficient tensor $\mathcal{W} \in \mathbb{R}^{p_1 \times \cdots \times p_M}$ measures how each variable of $\mathcal{X}$ contributes to the response. We estimate $\mathcal{W}$ with tensor regression\footnote{Based on the order of response, there are different types of tensor regression problems. For example, the tensor-on-tensor regression methods analyze tensor responses. In this paper, we focus on the tensor-on-scalar case where the response is a scalar value.} that solves
\begin{equation}\label{eq:tensor-regression}
    \widehat{\mathcal{W}} = \argmin_{\mathcal{W}}~\sum_{i=1}^N~\left(y_i - \langle \mathcal{W}, \mathcal{X}_i \rangle\right)^2.
\end{equation}

\subsection{Regularized Tensor Regression}
To reduce the model complexity, state-of-the-art tensor regression methods always add sparse or low-rank constraints on estimations and solve the regularized tensor regression
\begin{equation}\label{eq:regularized-tensor-regression}
    \widehat{\mathcal{W}} = \argmin_{\mathcal{W}}~\sum_{i=1}^N~\left(y_i - \langle \mathcal{W}, \mathcal{X}_i \rangle\right)^2~+~\mathcal{R}(\mathcal{W})
\end{equation}
with a regularization term $\mathcal{R}(\cdot): \mathbb{R}^{p_1 \times \cdots \times p_M} \mapsto \mathbb{R}$. Different regularizations lead to various structural properties.

\paragraph{Regularization for Sparsity} Tensor data is usually high-dimensional. Therefore, to solve the ill-defined tensor regression problem, some tensor regression models \citep{he2018boosted,zhou2013tensor} adopt ideas from sparse linear regression. Specifically, they assume only a small subset of variables contribute to the response, so $\mathcal{W}$ has many zero coefficients. They solve \eref{eq:regularized-tensor-regression} with an element-wise $\ell_1$ norm enforcing the coefficient sparsity as
\begin{equation}\label{eq:tensor-l1-norm}
    \begin{aligned}
    \mathcal{R}_{\text{sparse}}(\mathcal{W}) ~=~ ||\mathcal{W}||_{1}~=\sum_{i_1=1}^{p_1}\cdots\sum_{i_M = 1}^{p_M}~~\lvert \mathcal{W}_{i_1\cdots i_M} \rvert.
    \end{aligned}
\end{equation}
Our model uses the element-wise $\ell_1$ norm to achieve prediction sparsity.

\paragraph{Regularization for Low-Rankness} On the other hand, previous works \citep{song2017multilinear,li2019sturm,zhou2013tensor,he2018boosted} also use low-rank constraints to reduce model complexity. But computing tensor rank is NP-hard \citep{shitov2016hard,lim2009most}. So they usually optimize with tensor decomposition to compute the best fitting rank in a tractable way \citep{rabanser2017introduction}. CP decomposition and Tucker decomposition are two widely used tensor decomposition techniques \citep{tucker1966some,rabanser2017introduction}. But CP decomposition methods suffer the drawbacks of slow convergence and inaccurate prediction. Compared to CP decomposition, Tucker decomposition is a direct extension of singular value decomposition (SVD), hence, it better cooperates with existing optimization algorithms. So in this paper, we use Tucker decomposition as
\begin{equation}\label{eq:tucker-decomp}
    \mathcal{W} = \mathcal{C} \times_1 W_{(1)} \times_2 W_{(2)} \times_3 \cdots \times_M W_{(M)} 
\end{equation}
where $\mathcal{C} \in \mathbb{R}^{p_1 \times \cdots \times p_M}$ is the core tensor. Matrix $W_{(m)} \in \mathbb{R}^{p_m \times \prod_{k \neq m}p_k}$ is the result of unfolding the tensor $\mathcal{W}$ along the $m$-th order. $\times_m$ is the $m$-mode product operator defined in Section \ref{sec:notation}. The tensor $\mathcal{W}$ is low-rank as long as $\left\{ W_{(1)}, \cdots, W_{(M)}\right\}$ are all low-rank.

But it is difficult to directly obtain the tensor rank through \eref{eq:tucker-decomp}. Fortunately, as proven in \cite{tomioka2010estimation,liu2012tensor}, we can use a tensor nuclear norm regularization as the convex relaxation of the tensor rank. Tensor nuclear norm is the summation of $M$ matrix nuclear norm
\begin{equation}\label{eq:tensor-nuclear-norm}
    \begin{aligned}
        \mathcal{R}_{\text{low}}(\mathcal{W}) ~=~ ||\mathcal{W}||_{*}~=~\frac{1}{M}\sum\limits_{m=1}^{M}||W_{(m)}||_{*},
    \end{aligned}
\end{equation}
where $||W_{(m)}||_{*}$ is the matrix nuclear norm for the $m$-th mode. This tensor nuclear norm is an extension of a matrix nuclear norm and is proven to automatically obtain a low-rank tensor both accurately and reliably. So our model uses the tensor nuclear norm to reduce model complexity.


%% file: sec/method.tex
\begin{figure*}
    \centering
    \includegraphics[width=\textwidth]{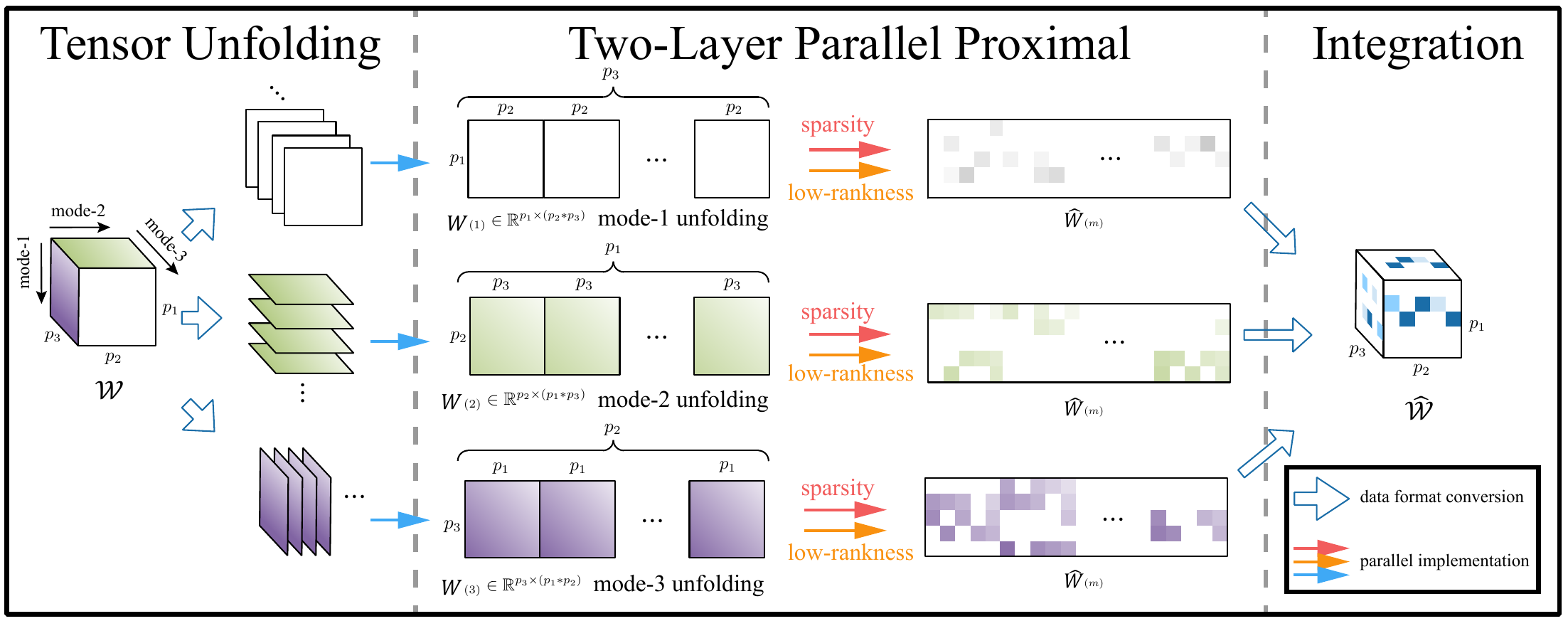}
    \caption{The basic idea of \NAME{}. We first unfold the tensor along each mode axis. Then for each mode, we parallelly estimate sparse and low-rank coefficients with proximal gradient descent. Finally, all mode estimations are integrated to obtain the final solution.}
    \label{fig:basic-idea}
\end{figure*}

\section{Method}\label{sec:method}

\paragraph{Model Overview} We propose the Sparse and Low-Rank High-Order Tensor Regression (\NAME{}) method, which optimizes a tensor regression problem with sparse and low-rank regularizers. We extend a linear regression framework to fit the tensor data and multiple regularizers. Also, to speed up the optimization, we propose a fast and scalable algorithm that computes predictions for each mode in parallel. \figref{fig:basic-idea} gives a schematic of our model.

\subsection{Optimization Problem: Multi-Regularized Tensor Regression with Fast Approximation}
How to analyze the relationship between vector variables and responses in high-dimensional settings has been well studied in the linear regression literature. Among various methods, Elem-Ridge \citep{yang2014elementary} is a novel estimation framework that enables fast optimization. Specifically, given variables $X \in \mathbb{R}^{N \times p}$ and responses $y \in \mathbb{R}^N$ of $N$ observations, Elem-Ridge assumes $y = Xw + \gamma$ and solves
\begin{equation}\label{eq:EE}
\begin{small}
  \begin{aligned}
    \widehat{w}~=~\argmin_{w \in \mathbb{R}^p}~\mathcal{R}(w) \qquad \text{s.t.}~~\mathcal{R}^*\left(w-\left(X^\top X+\varepsilon \mathbf{I}\right)^{-1}X^\top y\right) \leq \lambda.
  \end{aligned}
  \end{small}
\end{equation}
Here, $\mathcal{R}^*$ is the dual norm of $\mathcal{R}$,  $\mathbf{I}$ is an identity matrix, and $\varepsilon$ is a hyper-parameter dealing with the non-invertible sample covariance in high-dimensional cases. $\lambda$ is another hyper-parameter setting the constraint bound.

Elem-Ridge inherits the primal-dual idea of the well-known Dantzig selector \citep{candes2007dantzig} that obtains structural properties by minimizing the norm $\mathcal{R}$ while regressing the predictors $y$ through the dual-norm $\mathcal{R}^*$. For example, if one requires sparsity, Elem-Ridge should use $\ell_1$ norm as $\mathcal{R}$ for variable selection and uses $\ell_\infty$ as the dual. The advantage of Elem-Ridge is it starts the optimization of coefficients $w$ from an approximation $\left(X^\top X+\varepsilon \mathbf{I}\right)^{-1}X^\top y$ instead of a random feasible point. The approximation is a Ridge estimator, denoting it is not too distinct from the optimal solution but only lacks demanded structural properties. Therefore, the optimization can converge in a few iterations. Moreover, because the approximation only computes once and can be easily accelerated with GPUs, Elem-Ridge significantly reduces the time cost.

In this paper, we derive a novel tensor regression method \NAME{} by extending Elem-Ridge to the tensor data and integrating multiple regularizers. We use the element-wise norm and tensor nuclear norm defined in \eref{eq:tensor-l1-norm} and \eref{eq:tensor-nuclear-norm} to obtain sparse and low-rank estimations. Concretely, given $N$ samples of $M$-order tensor $\mathcal{X} \in \mathbb{R}^{N \times p_1 \times \cdots \times p_M}$ and corresponding responses $y \in \mathbb{R}^{N}$, our \NAME{} aims to solve 
\begin{equation}\label{eq:key-equation}
    \begin{aligned}  
    \widehat{\mathcal{W}}~=~\argmin_{\mathcal{W}}&~~\mathcal{R}_{\text{sparse}}(\mathcal{W}) ~+~ \mathcal{R}_{\text{low}}(\mathcal{W}) \\
      \text{s.t.}&~~\mathcal{R}^*_{\text{sparse}}\left( \mathcal{W}- \widetilde{\mathcal{W}} \right) ~\leq~ \lambda_{\text{sparse}}\\
      &~~\mathcal{R}^*_{\text{low}}\left( \mathcal{W}- \widetilde{\mathcal{W}} \right) ~\leq~ \lambda_{\text{low}}
    \end{aligned}
\end{equation}
with the approximation and dual norms as
\begin{equation}
    \begin{aligned}
    & \widetilde{\mathcal{W}} = \mathcal{T}_{\mathcal{P}}\left( \left(X^\top X + \varepsilon \mathbf{I}\right)^{-1}X^\top y \right) \\
    &\mathcal{R}^*_{\text{sparse}}\left( \mathcal{W}- \widetilde{\mathcal{W}} \right) = ||\mathcal{W}-\widetilde{\mathcal{W}}||_{\infty}\\
    &\mathcal{R}^*_{\text{low}}\left( \mathcal{W}- \widetilde{\mathcal{W}} \right) = ||\mathcal{W}-\widetilde{\mathcal{W}}||_{2}.
    \end{aligned}
\end{equation}
We extend the approximation in \eref{eq:EE} to tensors through unfolding-folding operations. The approximation $\widetilde{\mathcal{W}}$ is computed by first unfolding $\mathcal{X}$ into matrix $X$ through $X = \text{concat}(X_1,\cdots,X_N) \in \mathbb{R}^{N \times \prod_m p_m}$ with each row $X_i = \text{vec}(\mathcal{X}_i) \in \mathbb{R}^{\prod_m p_m}$ a vectorization of the $i-$th observation. Then we use the folding operation $\mathcal{T}_{\mathcal{P}}(\cdot)$ to convert the Ridge estimator back to the tensor, given the size of each mode as $\mathcal{P}= \{p_1, p_2, \cdots, p_M\}$. Because $\mathcal{R}_{\text{sparse}}$ is the element-wise $\ell_1$ norm, its dual is element-wise $\ell_\infty$ norm. Also, notice the tensor nuclear norm $\mathcal{R}_{\text{low}}$ is a summation of the matrix nuclear norm, its dual can be easily extended from the matrix spectral norm\footnote{We are not directly applying these norm functions on tensor data. Notice that both $\ell_1$ norm and tensor nuclear norm can be easily reformulated into the combination of multiple matrix norms, we can straightforwardly define them and corresponding dual norms on tensors. See \eref{eq:decompose-key-equation}}.

\subsection{Optimization Solution: Two-Layer Parallel Proximal Algorithm}
Based on the definition of element-wise $\ell_1$ norm, we can write is as $||\mathcal{W}||_1=\frac{1}{M}\sum\limits_{m=1}^{M}||W_{(m)}||_1$, the average of element-wise $\ell_1$ norm for each mode. This implies the $\ell_1$ norm computation for each mode $m$ is independent. Moreover, in \eref{eq:tensor-nuclear-norm}, the tensor nuclear norm is also defined as the average of each mode's nuclear norm. Therefore, we can decompose \eref{eq:key-equation} into $M$ parallel sub-tasks  
\begin{equation}\label{eq:decompose-key-equation}
    \begin{aligned}
      \widehat{W}_{(m)}=\argmin_{W_{(m)}}& \qquad ||W_{(m)}||_1 + ||W_{(m)}||_{*} \\
      \text{s.t.}& \qquad ||W_{(m)}-\widetilde{W}_{(m)}||_{\infty} \leq \lambda_{\text{sparse}}\\
      & \qquad ||W_{(m)}-\widetilde{W}_{(m)}||_{\text{2}} \leq \lambda_{\text{low}}.
    \end{aligned}
\end{equation}
After estimating $\widehat{W}_{(m)}$ for all modes $m=1, \cdots, M$, we integrate them into the final estimation 
\begin{equation}\label{eq:final-solution}
    \widehat{\mathcal{W}}=\frac{1}{M}\sum\limits_{m=1}^M \mathcal{T}_\mathcal{P} \left( \widehat{W}_{(m)} \right).
\end{equation}

We solve each sub-task (\eref{eq:decompose-key-equation}) through the parallel proximal algorithm \cite{combettes2011proximal}. We let $W \triangleq W_{(m)}$ for simplicity. Estimating $W_{(m)}$ is equivalent to
\begin{equation}\label{eq:proximal-problem}
    \begin{aligned}
    \argmin\limits_{W_1 = W_2 = W_3 = W_4}&~~f_1(W_1)+f_2(W_2)+f_3(W_3)+f_4(W_4)
    \end{aligned}
\end{equation}
by converting norms and corresponding constraints into 
\begin{equation}
    \begin{aligned}
        &f_1(W) ~=~ ||W||_1 \qquad\qquad\qquad\qquad\qquad  f_2(W) ~=~ ||W||_* \\
        &f_3(W) ~=~ \mathcal{I}_{\{||W-\widetilde{W}_{(m)}||_\infty \leq \lambda_{\text{sparse}} \}}(W) \qquad f_4(W) ~=~ \mathcal{I}_{\{||W-\widetilde{W}_{(m)}||_{2} \leq \lambda_{\text{low}} \}}(W)
    \end{aligned}
\end{equation}
where $\mathcal{I}_{C}(W)$ is an indicator function of set $C$ as $\mathcal{I}_{C}(W)=0$ if $W \in C$, otherwise $\mathcal{I}_{C}(W)=\infty$. 

The parallel proximal algorithm solves \eref{eq:proximal-problem} with proximal operators that compute convex approximations for non-differentiable functions like $\ell_1$ norm and indication function. Here, we define proximal operators for \eref{eq:proximal-problem}. Specifically, for $f_1$ and $f_3$, we have
\begin{equation}
    \begin{aligned}
        \text{prox}_{f_1}(W; \lambda_{\text{sparse}}) = S_{\lambda_{\text{sparse}}}(W) \quad\text{with}\quad  \left[ S_{\lambda_{\text{sparse}}}(W) \right]_{ij} = \sign(W_{ij})\max\left\{ \lvert W_{ij} \rvert - \lambda_{\text{sparse}},~0 \right\}
    \end{aligned}
\end{equation}
and
\begin{equation}
    \text{prox}_{f_3}(W; \lambda_{\text{sparse}}) = 
    \left\{
      \begin{aligned}
        \widetilde{W}_{ij}, &~~~~\lvert W_{ij} - \widetilde{W}_{ij} \rvert \leq \lambda_{\text{sparse}} \\
        \widetilde{W}_{ij} + \lambda_{\text{sparse}}, &~~~W_{ij} - \widetilde{W}_{ij} > \lambda_{\text{sparse}} \\
        \widetilde{W}_{ij} - \lambda_{\text{sparse}}, &~~~W_{ij} - \widetilde{W}_{ij} < -\lambda_{\text{sparse}} 
      \end{aligned}
    \right. .
\end{equation}
The other two nuclear-norm-relevant proximal operators are computed based on singular value decomposition (SVD). For a matrix $W$, SVD decomposes it into $W=U \Sigma V^\top$ in which $\Sigma$ is a diagonal matrix with singular values on its diagonal and $U$, $V$ are left and right singular vectors correspondingly. Based on this, we have
\begin{equation}
    \text{prox}_{f_2}(W; \lambda_{\text{low}}) = US_{\lambda_{\text{low}}}(\Sigma)V^\top \quad\text{with}\quad  \left[ S_{\lambda_{\text{low}}}(\Sigma) \right]_{ii} = \sign(\Sigma_{ii})\max\left\{ \lvert \Sigma_{ii} \rvert - \lambda_{\text{low}},~0 \right\}
\end{equation}
and
\begin{equation}
    \text{prox}_{f_4}(W; \lambda_{\text{low}}) = 
    \left\{
      \begin{aligned}
        \widetilde{W}, &~~~~ \sigma_{\text{max}}(W) \leq \lambda_{\text{low}} \\
        US_{\lambda_{\text{low}}}(\Sigma)V^\top + \widetilde{W},&~~~\sigma_{\text{max}}(W) > \lambda_{\text{low}} 
      \end{aligned}
    \right.
\end{equation}
where $\sigma_{\text{max}}(W)$ represents the maximum eigenvalue of $W$. Given these proximal operators, our optimization solution is summarized in Algorithm \ref{algo:pseudo}. Notice the sub-task optimizations are parallel, and the computation of four proximal operators within each sub-task is also parallel. Hence, our algorithm obtains the solution in the manner of two-layer parallelism.

\begin{algorithm}[!htb]
    \caption{~~Parallel Proximal Based Algorithm for \NAME{}}
    \label{algo:pseudo}
    \begin{algorithmic}[1]
      \STATE {\bfseries Input:} $\mathcal{X} \in \mathbb{R}^{N \times p_1 \times p_2 \times \cdots \times p_M}$, $y \in \mathbb{R}^{N}$, initial approximation $\widetilde{\mathcal{W}}$, the maximum number of iterations $T$, learning rate $\rho \in [0,2]$, and tuning parameters $\mathbf{c}=(\lambda_{\text{sparse}}, \lambda_{\text{sparse}}, \lambda_{\text{low}}, \lambda_{\text{low}})$.  
      \vspace{0.1in}
      \FOR{$m=1$ to $M$ \textbf{parallelly}}
        \STATE \textbf{Initialize} $W_{(m)} = W^1_{(m)1} = W^1_{(m)2}  = W^1_{(m)3} = W^1_{(m)4} = \widetilde{W}_{(m)}$
        \FOR{$t=1$ to $T$}
          \FOR{$i=1,2,3,4$ \textbf{parallelly}}
            \STATE $a_i^t=\text{prox}_{f_i}\left( W_{(m)i}^t~;~4\mathbf{c}_i \right)$
          \ENDFOR
          \STATE $a^t = \frac{1}{4}\sum\limits_{i=1}^4 a_i^t$
          \FOR{$i=1,2,3,4$}
            \STATE $W_{(m)i}^{t+1}=W_{(m)i}^{t}+\rho(2a^t-W_{(m)}-a_i^t)$
          \ENDFOR
          \STATE $W_{(m)} = W_{(m)} + \rho(a^t-W_{(m)})$
        \ENDFOR
        \STATE $\widehat{\mathcal{W}}_m = \mathcal{T}_\mathcal{P} \left( W_{(m)} \right)$
      \ENDFOR
      \STATE{\bfseries Output:}  $\widehat{\mathcal{W}} = \frac{1}{M}\sum\limits_{m=1}^M\widehat{\mathcal{W}}_m$.
    \end{algorithmic}
\end{algorithm}

\subsection{Running Time Complexity}\label{sec:time}

First of all, the initial approximation $\widetilde{\mathcal{W}}$ is only computed once and reused in the algorithm. Because computing $\widetilde{\mathcal{W}}$ requires only simple operations such as matrix multiplication and matrix inversion, this part can be easily speeded up with multi-thread computing and rapidly obtained as the pre-condition of our algorithm. In Sec. \ref{sec:exp}, we show that the individual pre-condition calculation of \NAME{} enables more rapid estimation with hyper-parameter tuning. Given $\widetilde{\mathcal{W}}$, totally $M$ sub-tasks are solved simultaneously using parallel proximal-based algorithm Algorithm \ref{algo:pseudo}. Within each sub-task, the computation is dominated by the SVD procedure with $O(p_m(\prod_{k \neq m} p_k)^2)$ time complexity. So overall, by virtue of our two-layer parallel solution, the computational bottleneck of solving \NAME{} can be considered as only $O(\max\limits_{m}\{p_m(\prod_{k \neq m}^M p_k)^2\})$.




%% file: sec/theory.tex
\section{Theoretical Analysis}\label{sec:theorem}
We now prove the convergence rate of \NAME{}. We follow the proof of \cite{yang2014elementary} and assume the following:

\paragraph{(C1: sparse)} The optimal coefficient $\mathcal{W}^*$ has exactly $k$ non-zero elements. 

\paragraph{(C2: low-rank)} The optimal coefficient $\mathcal{W}^*$ is an $R$-rank tensor, where $R = \max\limits_{\mathcal{A} \in \mathbb{R}^{p_1 \times \cdots \times p_M}}(r_{\perp}(\mathcal{A}))$ and $r_{\perp}(\mathcal{A})$ denotes the orthogonal rank of $\mathcal{A}$. The orthogonal rank is the smallest number that satisfies $\mathcal{A}=\sum_{r=1}^{r_{\perp}(\mathcal{A})}\mathcal{U}_r$ with $<\mathcal{U}_{r_1}, \mathcal{U}_{r_2}>=0,~r_1 \neq r_2$ for $1 \leq r_1 \leq r_{\perp}(\mathcal{A}),~1 \leq r_2 \leq r_{\perp}(\mathcal{A})$.

\begin{theorem}\label{theorem:error-bound}
    Suppose we solve \eref{eq:key-equation} with proper controlling parameters $\lambda_{\text{sparse}}$ and $\lambda_{\text{low}}$. Then, the estimation satisfies the error bound
    \begin{equation}
        ||\widehat{\mathcal{W}}-\mathcal{W}^*||_F \leq 4\sqrt{2}\left(\lambda_{\text{sparse}} \sqrt{\prod\limits_{m=1}^{M}p_m} + \lambda_{\text{low}}\sqrt{R}\right).
    \end{equation}
\end{theorem}

\begin{corollary}\label{corollary:error-bound-3D}
    In the three-order tensor case where $\mathcal{W} \in \mathbb{R}^{p_1 \times p_2 \times p_3}$, the estimation of \eref{eq:key-equation} satisfies the error bound
    \begin{equation}
        ||\widehat{\mathcal{W}}-\mathcal{W}^*||_F \leq 4\sqrt{2}\left(\lambda_{\text{sparse}} \sqrt{\prod\limits_{m=1}^{M}p_m} + \lambda_{\text{low}} \max\limits_{k = 1,2,3}\{R^\prime_k\}\right),
    \end{equation}
    where $r_m = \text{rank} \left( \mathcal{W}_{(m)} \right)$ denotes the rank of the unfolded matrix. $R^\prime_1 = \sqrt{r_1\min\{r_2,r_3\}}$, $R^\prime_2 = \sqrt{r_2\min\{r_1,r_3\}}$, and $R^\prime_3 = \sqrt{r_3\min\{r_1,r_2\}}$.
\end{corollary}

All proofs are provided in the appendix.



%% file: sec/related_work.tex
\section{Related Works}

Some tensor regression methods \citep{he2018boosted,zhou2013tensor,guo2011tensor} have been proposed based on CP decomposition. Generally, these methods aim at inferring decomposed components to approximate low-rank estimations. For example, \cite{zhou2013tensor} proposed Generalized Linear Tensor Regression Model using the generalized linear model (GLM). In addition, \cite{he2018boosted} recently proposed Stagewise Unit-Rank Tensor Factorization (SURF) exploiting the divide-and-conquer strategy where the sub-task has a similar formulation of Elastic Net \citep{zou2005regularization}. Almost all the CP-decomposition-based methods require prior knowledge of the CP-rank $R$. However, we always have little information about it in real-world applications. Even if we can use techniques, such as cross-validation, to select $R$ from a wide range, choosing the $R$ value becomes complicated and computationally expensive for large-scale data. Moreover, the larger $R$ is, the more computational time is required for these methods. Therefore, these methods are not suitable for real-world applications.

In another line of work, structural constraints are directly applied to the coefficient tensor rather than its decomposed components in order to avoid expensive decomposition. For instance, in Regularized multilinear regression and selection (Remurs) \citep{song2017multilinear}, the tensor nuclear norm and $\ell_1$ norm are used. However, these methods are computationally expensive because non-differential regularizers exist in their objective function and the lack of parallelism. We compare our \NAME{} with state-of-the-art tensor regression models in \tabref{tab:bottleneck}. \NAME{} outperforms other methods on model abilities and computational time complexity.

\begin{table}[!htb]
    \centering
    \caption{Comparison between \NAME{} and other tensor regression models. $T$ denotes the number of iterations for iterative method, $N$ is the number of samples, $M$ is the number of modes, and $R$ is the CP-rank. $\mathbf{P}=\prod_{m=1}^Mp_m$ and $\mathbf{P_{\backslash m}}=\prod_{k \neq m}^M p_k$. $T$ is the number of SURF iterations. We compare their computational bottlenecks and properties.}
    \label{tab:bottleneck}
    \resizebox{\textwidth}{!}{%
    \begin{tabular}{|c||c|c|c|c|c|c|}
    \Xhline{1.6pt}
     & \NAME & Remurs & GLTRM & orTRR & SURF & LR \\ \Xhline{0.8pt}
    Comp. Bottleneck & \begin{tabular}[c]{@{}c@{}}$O(\max\limits_{m}\{p_m \cdot \mathbf{P^2_{\backslash m}}\})$. \end{tabular} & \begin{tabular}[c]{@{}c@{}}$O(\sum\limits_{m=1}^M\{p_m \cdot \mathbf{P^2_{\backslash m}}\})$\end{tabular} & \begin{tabular}[c]{@{}c@{}}$O(R \sum\limits_{m=1}^Mp_m^3)$ \end{tabular} & $O(M \cdot \mathbf{P^3})$ & $O(TN \cdot \sum\limits_{m=1}^M \mathbf{P_{\backslash m}})$ & \begin{tabular}[c]{@{}c@{}}$O(N \cdot \mathbf{P^2})$\end{tabular} \\ \Xhline{0.2pt}
    Auto-Explored Rank & $\checkmark$ & $\checkmark$ & $\times$ & $\checkmark$ & $\times$ & $\times$\\ \Xhline{0.2pt}
    Sufficient Sparsity & $\checkmark$ & $\checkmark$ & $\checkmark$ & $\times$ & $\checkmark$ & $\checkmark$\\ \Xhline{0.2pt}
    Structure Reserved & $\checkmark$ & $\checkmark$ & $\checkmark$ & $\checkmark$ & $\checkmark$ & $\times$ \\ \Xhline{1.6pt}
    \end{tabular}
    }
    \vspace{-8pt}
  \end{table}

%% file: sec/result.tex
\section{Experiment}\label{sec:exp}

\paragraph{Baselines} We compare our \NAME{} with four previously proposed methods, representing different groups of tensor regression methods, including (1) Linear regression models, specifically, Lasso and Elastic Net (with trade-off ratio between $\ell_1$ and $\ell_2$ norm being 0.5)
\footnote{Here, we employ Lasso and Elastic Net on the vectorized data.}
, (2) Remurs \citep{song2017multilinear}, and (3) SURF \citep{he2018boosted}. All the methods are implemented in {\tt MATLAB}. 

\paragraph{Evaluation metrics} For experiments on simulated datasets (\secref{sec:exp-sim}), we report the computational time cost (in seconds) and prediction mean squared error (MSE) for all the methods. For experiments on the video action recognition dataset (\secref{sec:exp-real}), we report the computational time cost (in seconds) and area under receiver operating characteristic (AUROC) for each pair of action labels.

\paragraph{Hyper-arameter tuning} Tuning hyper-parameters of all the methods are selected through cross-validation procedures which take the average performance on validation datasets as the selecting criteria. We tune hyper-parameters from a wide range of values to ensure each method achieves its best performance. The ranges of hyper-parameters are listed in the appendix. 

\paragraph{Other setups} We set the maximal number of iterations to be $1000$ for all the methods and let them terminate when the iteration update $\frac{||\mathcal{W}^{t+1}-\mathcal{W}^t||_F}{||\mathcal{W}^t||_F} \leq 10^{-4}$. We run every single experiment ten times and report the average value of metrics over these ten trials.

\subsection{Simulated Data: Tensor Regression}\label{sec:exp-sim}
We first test our model on simulated datasets. The dataset is generated through the following steps: 
\begin{itemize}
    \item \textbf{(Step 1)} Specify the optimal coefficient tensor $\mathcal{W}^* \in \mathbb{R}^{p_1 \times p_2 \times \cdots \times p_M}$ and $N$ samples $\mathcal{X} \in \mathbb{R}^{N \times p_1 \times p_2 \times \cdots \times p_M}$ with each element drawn from the normal distribution $\mathcal{N}(0,~1)$. 
    
    \item \textbf{(Step 2)} Randomly set $s\%$ elements of $\mathcal{W}$ to be $0$.
    
    \item \textbf{(Step 3)} Compute $N$ responses $y \in \mathbb{R}^{N}$ through $y_i = \langle \mathcal{W}^*,\mathcal{X}_i \rangle + 0.1\gamma_i$, where the noise $\varepsilon_i$ is generated from the normal distribution $\mathcal{N}(0,~0.1)$.
\end{itemize}
We simulate 3D and 4D datasets with different shapes by fixing the sparsity level $s\%=80\%$. Because the SURF implementation does not apply to 4D data, we omit it in 4D data experiments. This also indicates its limitations in broader applications.

We report MSE values of estimation on high-dimensional simulated datasets in \tabref{tab:MSE}. The number of samples is determined through $N=8\% \cdot \prod_{m}^{M}p_m$ for each dataset in order to construct high-dimensional settings. The result indicates that \NAME{} has the best estimation in most cases, while its estimations are only slightly worse than the best in other cases. The MSE values of tensor regression models are significantly lower than those of linear regression models, indicating that linear regression indeed discards the structural information of tensors. In addition, we investigate how estimations improve if we use more samples. Specifically, we simulate 3D datasets with the shape of $20 \times 20 \times 5$ and vary the number of samples $N$ from 50 to 400. \tabref{tab:high-dimensional} shows that \NAME{} obtains the best estimation under almost all conditions. Moreover, the MSE decreases when more samples are provided, as we expect.

\begin{table}[!htp]
  \centering
  \caption {MSE on simulated datasets of different variable sizes. The bold number denotes the best method and the underlined value represents the second best result.}
  \label{tab:MSE}
  \resizebox{.6\columnwidth}{!}{%
  \begin{tabular}{cccccc}
  \Xhline{1.2pt}
  size & \NAME{}{} & Remurs & SURF & Lasso & Elastic Net \\ 
  \cline{1-6}
  \multicolumn{6}{c}{\emph{3D Data} -- 8\% samples}\\
  30 $\times$ 30 $\times$ 5 & \textbf{0.9186} & {\ul 0.9190} & 0.9289 & 1.9381 & 1.9377 \\
  35 $\times$ 35 $\times$ 5 & \textbf{0.9336} & {\ul 0.9370} & 0.9527 & 2.0147 & 2.0147 \\
  40 $\times$ 40 $\times$ 5 & \underline{0.9073} & \textbf{0.9072} & 1.0006 & 2.1059 & 2.1065 \\
  \cdashline{1-6}[0.8pt/2pt]
  \multicolumn{6}{c}{\emph{4D Data} -- 8\% samples}\\
  20 $\times$ 20 $\times$10 $\times$ 5 & \textbf{0.9150} & {\ul 0.9177} & \multirow{3}{*}{N/A} & 2.1388 & 2.1373\\
  25 $\times$ 25 $\times$10 $\times$ 5 & \textbf{0.9071} & {\ul 0.9101} & & 1.9696 & 1.9696\\
  30 $\times$ 30 $\times$10 $\times$ 5 & \textbf{0.9110} & {\ul 0.9123} & & 1.9754 & 1.9745\\

  \Xhline{1.2pt}
  \end{tabular}
  }
\end{table}

\begin{table}[!htp]
  \centering
  \caption {MSE on simulated datasets of different numbers of samples. The number of samples varies from 50 to 400. The bold number denotes the best method and the underlined value represents the second best result.}
  \label{tab:high-dimensional}
  \resizebox{.6\columnwidth}{!}{%
  \begin{tabular}{ccccccccccccccc}
  \Xhline{1.2pt}
  sample numbers ($N$) & \NAME{} & Remurs & SURF & Lasso & Elastic Net \\
  \cline{1-6}
  50 & \textbf{1.6123} & {\ul 1.6198} & 1.6439 &4.5083 & 4.5759 \\
  100 & \textbf{1.0798} & {\ul 1.0946} & 1.7101 &1.6433 & 1.6433 \\
  150 & \underline{0.9295} & \textbf{0.9190} & 0.9953 &1.6777 & 1.6502 \\
  200 & \textbf{0.8351} & 0.8469 & {\ul 0.8376} & 1.9072 & 0.8376 \\
  250 & \textbf{0.7130} & 0.7267 & {\ul 0.7199} & 1.3757 & 1.3708 \\
  300 & \textbf{0.7282} & {\ul 0.7325} & 0.7524 & 1.7316 & 1.6938 \\
  350 & \textbf{0.6275} & {\ul 0.6275} & 0.6379 & 1.3207 & 1.2804 \\ 
  400 & \textbf{0.5954} & {\ul 0.5969} & 0.5975 & 1.1487 & 1.1450 \\
  \Xhline{1.2pt}
  \end{tabular}
  }
\end{table}

The benefit of our \NAME{} is it uses a fast approximation to speed up computations. Especially in hyper-parameter tuning, \NAME{} only needs to compute the approximation $\widetilde{\mathcal{W}}$ once (for one $\epsilon$ value) when changing the value of controlling hyper-parameters. Since hyper-parameter tuning is a necessary procedure in model selection, fast approximation enables \NAME{} to obtain the best solution with much fewer time costs than previous methods. To validate this, we record the time cost of hyper-parameter tuning for \NAME{} and Remurs. They have two common hyper-parameters $\lambda_{\text{sparse}}$ and $\lambda_{\text{low}}$ controlling the degree of sparsity and low-rankness. We tune each of them from seven values $\{0.005, 0.01, 0.05, 0.1, 0.5, 1, 5\}$. Moreover, \NAME{} has an extra hyper-parameter $\epsilon$ for approximation computation, which we tune from three values $\{0.1, 0.2, 0.3\}$. So \NAME{} and Remurs use 147 and 49 groups of hyper-parameters, respectively. We record the time of the method running with each hyper-parameter group and report the total time cost on all groups. Table \tabref{tab:cross_validation_simulated} indicates that \NAME{} is faster than Remurs by orders of magnitude, and the speedup becomes increasingly evident as the size of the data increases. Therefore, even though our \NAME{} has one more hyper-parameter than Remurs, its time cost is largely reduced by virtue of the fast and one-time approximation calculation.

\begin{table}[!htp]
\centering
  \caption {Time costs (in seconds) of hyper-parameter tuning. The ``speedup'' is computed by dividing Remurs cost by \NAME{} cost.}
  \label{tab:cross_validation_simulated}
   \resizebox{0.6\textwidth}{!}{
\begin{tabular}{cccc}
\Xhline{1.2pt}
size             & SLTR      & Remurs & speedup     \\ \cline{1-4}
\multicolumn{4}{c}{\emph{3D Data} -- 8\% samples}  \\
30 × 30 × 5      & 16.793568 & 510.050226  & \textbf{30.37}$\times$ \\
30 × 30 × 5      & 20.212862 & 602.292492  & \textbf{29.80}$\times$ \\
30 × 30 × 5      & 6.509343  & 684.737448  & \textbf{105.20}$\times$ \\
\cdashline{1-4}[0.8pt/2pt]
\multicolumn{4}{c}{\emph{4D Data} -- 8\% samples}  \\
20 × 20 × 10 × 5 & 10.975796 & 1474.090235 & \textbf{134.30}$\times$ \\
25 × 25 × 10 × 5 & 16.936689 & 3060.085998 & \textbf{180.67}$\times$ \\
30 × 30 × 10 × 5 & 23.276677 & 3394.478247 & \textbf{145.83}$\times$ \\ \hline
\end{tabular}
}
\end{table}

Overall, experiments on simulated datasets validate that \NAME{} predict better estimations with much less time costs.

\subsection{Real-World Case: Video Action Recognition}\label{sec:exp-real}

We then evaluate our method on the UCF101 \citep{soomro2012ucf101}, a video action recognition dataset. It collects 13320 videos of 101 action categories from YouTube. Each video has a different time length, ranging from less than 2 seconds to longer than 10 seconds, with each frame having a resolution of 320 \emph{pixels} $\times$ 240 \emph{pixels}. We focus on binary classification tasks and choose three pairs of categories: ``ApplyEyeMakeup'' vs. ``ApplyLipstick'', ``BaseballPitch'' vs. ``Basketball'', and ``BodyWeightSquats'' vs. ``Bowling''. We uniformly extract 15 frames with a fixed interval from each video and transform them into the grey scale. For each frame, we resize it into 32 \emph{pixels} $\times$ 24 \emph{pixels} by averaging neighboring pixels. So each sample is a $15 \times 32 \times 24$  tensor. For each pair of categories, we select $80\%$ of samples for training and the other 20\% for testing. We use AUROC scores to evaluate the classification performance.

\tabref{tab:UCF101} shows that \NAME{} reaches nearly the best AUROC scores. Only the AUROC scores of Remurs are comparable to \NAME{}. In the first pair of labels, \NAME{} has a slightly smaller AUROC value and performs much better than SURF and two linear models. In the other two cases, our \NAME{} has the most accurate estimations among all methods. Furthermore, we compare the hyper-parameter tuning time cost of \NAME{} (with 80 groups of hyper-parameters) and Remurs (with 49 groups of hyper-parameters). \tabref{tab:cross_validation_UCF101} shows that \NAME{} has a significant time advantage over Remurs. This implies that \NAME{} has good estimations with lesser time costs.

\begin{table*}[htb]
  \centering
  \caption {AUROC values on the UCF101 dataset. The bold number denotes the best AUROC value, while underlined number highlights the second best.}
  \label{tab:UCF101}
  \resizebox{.9\linewidth}{!}{%
  \begin{tabular}{cccccc}
  \Xhline{1.2pt}
  Label Pair & \NAME{} & Remurs & SURF & Lasso & Elastic Net \\
  ``ApplyEyeMakeup'' \& ``ApplyLipstick'' & \underline{0.931953} & \textbf{0.945266} & 0.64053 &  0.88006 & 0.887556\\
  ``BaseballPitch'' \& ``Basketball'' & \textbf{0.995074} & {\ul 0.995074} & 0.78695 & 0.964194 & 0.965473 \\
  ``BodyWeightSquats'' \& ``Bowling'' & \textbf{0.97756} & \underline{0.946704} & 0.32258 &  0.919753 & 0.930556 \\  \hline
  \Xhline{1.2pt}
  \end{tabular}
  }
\end{table*}

\begin{table}[!htp]
\centering
  \caption {Hyper-parameter tuning time cost (in seconds) on the UCF101 dataset.}
  \label{tab:cross_validation_UCF101}
     \resizebox{0.8\textwidth}{!}{
\begin{tabular}{cccc}
\hline
Lable pair                                & SLTR           & Remurs        & speedup \\ 
``ApplyEyeMakeup'' \& ``ApplyLipstick''     & 6.889112525    & 692.231181    &  \textbf{100.48}$\times$ \\
``BaseballPitch'' \& ``Basketball''         & 6.865089138    & 678.267903    &  \textbf{98.80}$\times$ \\
``BodyWeightSquats'' \& ``Bowling''         & 6.638173225    & 701.628673    &  \textbf{105.67}$\times$ \\ \hline

\end{tabular}}
\end{table}

\begin{figure*}[!htb]
    \centering
    \includegraphics[width=0.85\linewidth]{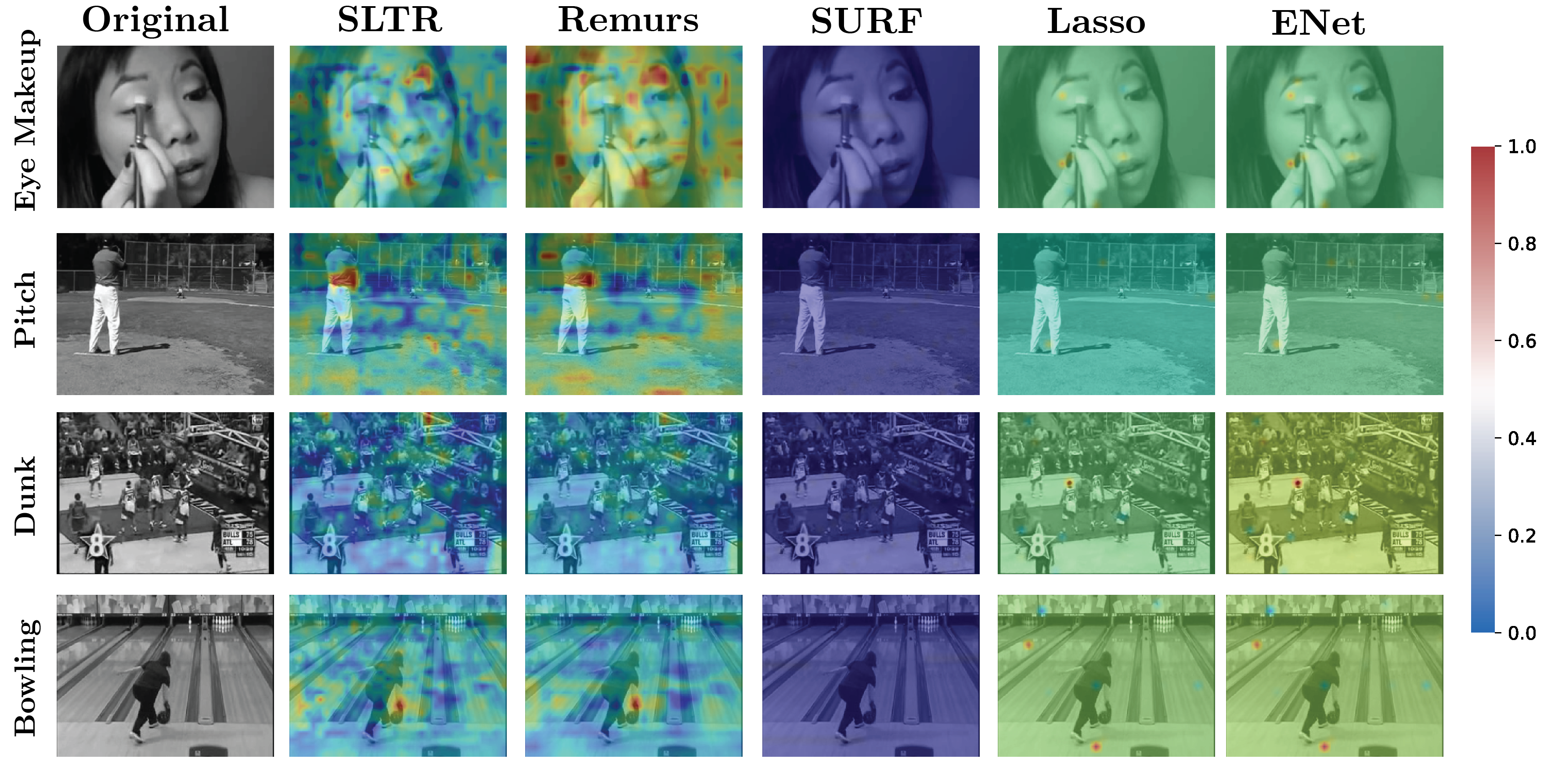}
    \caption{Example heatmaps of estimated coefficients on one frame of four videos.}
    \label{fig:video-heatmap}
\end{figure*}

Lastly, we visualize estimated coefficients in \figref{fig:video-heatmap}. The heatmap shows the superior interpretability of \NAME{}. For example, in a video of the ``ApplyEyeMakeup'' class (first row in \figref{fig:video-heatmap}), the focus of \NAME{} (i.e., high estimated weights) is mainly on the eyes. Remurs shows similar interpretations while \NAME{} estimations are more sparse. SURF estimations are all close to zeros, which explains its terrible AUROC value. Linear methods only focus on a few spots that cannot help with interpretations. Experiments on this classification task indicate that \NAME{} can give accurate and interpretable solutions with much fewer time costs.

%% file: sec/discussion.tex
\section{Conclusion}
This paper proposes a fast and scalable tensor regression method that directly imposes structural constraints on the tensor variables. A two-layer parallel solution is also proposed to solve the problem effectively. The benefit of our model is that fast approximation vastly reduces computational time costs, especially in hyper-parameter tuning.

The work may attract the interest of researchers from various fields when analyzing relationships for tensor data. The targeted areas include neuroscience and bioinformatics. For example, in neuroscience, the model can be applied to high-dimensional fMRI data to analyze the relationship between brain activities and disease symptoms.

In future work, we will explore variations of our method on data with modes denoting special relationships. For example, the \emph{time} mode of video data should have temporal dependencies. Adding extra constraints over such a relationship will increase the performance of \NAME{} in specific video analysis applications. Moreover, we will extend our model to more general cases regarding the tensor-on-vector or tensor-on-tensor regression.

%% file: main.bbl
\begin{thebibliography}{}

\bibitem[Ahmed et~al., 2020]{ahmed2020tensor}
Ahmed, T., Raja, H., and Bajwa, W.~U. (2020).
\newblock Tensor regression using low-rank and sparse tucker decompositions.
\newblock {\em SIAM Journal on Mathematics of Data Science}, 2(4):944--966.

\bibitem[Bhargava et~al., 2015]{bhargava2015and}
Bhargava, P., Phan, T., Zhou, J., and Lee, J. (2015).
\newblock Who, what, when, and where: Multi-dimensional collaborative
  recommendations using tensor factorization on sparse user-generated data.
\newblock In {\em Proceedings of the 24th international conference on world
  wide web}, pages 130--140. International World Wide Web Conferences Steering
  Committee.

\bibitem[Candes et~al., 2007]{candes2007dantzig}
Candes, E., Tao, T., et~al. (2007).
\newblock The dantzig selector: Statistical estimation when p is much larger
  than n.
\newblock {\em The annals of Statistics}, 35(6):2313--2351.

\bibitem[Cichocki et~al., 2016]{cichocki2016tensor}
Cichocki, A., Lee, N., Oseledets, I., Phan, A.-H., Zhao, Q., Mandic, D.~P.,
  et~al. (2016).
\newblock Tensor networks for dimensionality reduction and large-scale
  optimization: Part 1 low-rank tensor decompositions.
\newblock {\em Foundations and Trends{\textregistered} in Machine Learning},
  9(4-5):249--429.

\bibitem[Combettes and Pesquet, 2011]{combettes2011proximal}
Combettes, P.~L. and Pesquet, J.-C. (2011).
\newblock Proximal splitting methods in signal processing.
\newblock In {\em Fixed-point algorithms for inverse problems in science and
  engineering}, pages 185--212. Springer.

\bibitem[Guo et~al., 2011]{guo2011tensor}
Guo, W., Kotsia, I., and Patras, I. (2011).
\newblock Tensor learning for regression.
\newblock {\em IEEE Transactions on Image Processing}, 21(2):816--827.

\bibitem[He et~al., 2018]{he2018boosted}
He, L., Chen, K., Xu, W., Zhou, J., and Wang, F. (2018).
\newblock Boosted sparse and low-rank tensor regression.
\newblock In {\em Advances in Neural Information Processing Systems}, pages
  1009--1018.

\bibitem[Heinze et~al., 2018]{heinze2018variable}
Heinze, G., Wallisch, C., and Dunkler, D. (2018).
\newblock Variable selection--a review and recommendations for the practicing
  statistician.
\newblock {\em Biometrical journal}, 60(3):431--449.

\bibitem[Ji et~al., 2019]{ji2019survey}
Ji, Y., Wang, Q., Li, X., and Liu, J. (2019).
\newblock A survey on tensor techniques and applications in machine learning.
\newblock {\em IEEE Access}, 7:162950--162990.

\bibitem[Li et~al., 2013]{li2013some}
Li, N., Kindermann, S., and Navasca, C. (2013).
\newblock Some convergence results on the regularized alternating least-squares
  method for tensor decomposition.
\newblock {\em Linear Algebra and its Applications}, 438(2):796--812.

\bibitem[Li et~al., 2019]{li2019sturm}
Li, W., Lou, J., Zhou, S., and Lu, H. (2019).
\newblock Sturm: Sparse tubal-regularized multilinear regression for fmri.
\newblock In {\em International Workshop on Machine Learning in Medical
  Imaging}, pages 256--264. Springer.

\bibitem[Lim and Hillar, 2009]{lim2009most}
Lim, L.-H. and Hillar, C. (2009).
\newblock Most tensor problems are np hard.
\newblock {\em University of California, Berkeley}.

\bibitem[Liu et~al., 2012]{liu2012tensor}
Liu, J., Musialski, P., Wonka, P., and Ye, J. (2012).
\newblock Tensor completion for estimating missing values in visual data.
\newblock {\em IEEE transactions on pattern analysis and machine intelligence},
  35(1):208--220.

\bibitem[Lui, 2012]{lui2012least}
Lui, Y.~M. (2012).
\newblock A least squares regression framework on manifolds and its application
  to gesture recognition.
\newblock In {\em 2012 IEEE Computer Society Conference on Computer Vision and
  Pattern Recognition Workshops}, pages 13--18. IEEE.

\bibitem[Mitchell et~al., 2008]{mitchell2008predicting}
Mitchell, T.~M., Shinkareva, S.~V., Carlson, A., Chang, K.-M., Malave, V.~L.,
  Mason, R.~A., and Just, M.~A. (2008).
\newblock Predicting human brain activity associated with the meanings of
  nouns.
\newblock {\em science}, 320(5880):1191--1195.

\bibitem[Noroozi and Rezghi, 2020]{noroozi2020tensor}
Noroozi, A. and Rezghi, M. (2020).
\newblock A tensor-based framework for rs-fmri classification and functional
  connectivity construction.
\newblock {\em Frontiers in neuroinformatics}, 14:581897.

\bibitem[Rabanser et~al., 2017]{rabanser2017introduction}
Rabanser, S., Shchur, O., and G{\"u}nnemann, S. (2017).
\newblock Introduction to tensor decompositions and their applications in
  machine learning.
\newblock {\em arXiv preprint arXiv:1711.10781}.

\bibitem[Sharma and Gera, 2013]{sharma2013survey}
Sharma, L. and Gera, A. (2013).
\newblock A survey of recommendation system: Research challenges.
\newblock {\em International Journal of Engineering Trends and Technology
  (IJETT)}, 4(5):1989--1992.

\bibitem[Shitov, 2016]{shitov2016hard}
Shitov, Y. (2016).
\newblock How hard is the tensor rank?
\newblock {\em arXiv preprint arXiv:1611.01559}.

\bibitem[Song and Lu, 2017]{song2017multilinear}
Song, X. and Lu, H. (2017).
\newblock Multilinear regression for embedded feature selection with
  application to fmri analysis.
\newblock In {\em Thirty-First AAAI Conference on Artificial Intelligence}.

\bibitem[Soomro et~al., 2012]{soomro2012ucf101}
Soomro, K., Zamir, A.~R., and Shah, M. (2012).
\newblock Ucf101: A dataset of 101 human actions classes from videos in the
  wild.

\bibitem[Tomioka et~al., 2010]{tomioka2010estimation}
Tomioka, R., Hayashi, K., and Kashima, H. (2010).
\newblock Estimation of low-rank tensors via convex optimization.
\newblock {\em arXiv preprint arXiv:1010.0789}.

\bibitem[Tucker, 1966]{tucker1966some}
Tucker, L.~R. (1966).
\newblock Some mathematical notes on three-mode factor analysis.
\newblock {\em Psychometrika}, 31(3):279--311.

\bibitem[Wu and Lai, 2010]{wu2010tensor}
Wu, X. and Lai, J. (2010).
\newblock Tensor-based projection using ridge regression and its application to
  action classification.
\newblock {\em IET image processing}, 4(6):486--493.

\bibitem[Yang et~al., 2014]{yang2014elementary}
Yang, E., Lozano, A., and Ravikumar, P. (2014).
\newblock Elementary estimators for high-dimensional linear regression.
\newblock In {\em International Conference on Machine Learning}, pages
  388--396.

\bibitem[Yang et~al., 2017]{yang2017tensor}
Yang, Y., Krompass, D., and Tresp, V. (2017).
\newblock Tensor-train recurrent neural networks for video classification.
\newblock In {\em International Conference on Machine Learning}, pages
  3891--3900. PMLR.

\bibitem[Zhou et~al., 2013]{zhou2013tensor}
Zhou, H., Li, L., and Zhu, H. (2013).
\newblock Tensor regression with applications in neuroimaging data analysis.
\newblock {\em Journal of the American Statistical Association},
  108(502):540--552.

\bibitem[Zhu et~al., 2014]{zhu2014fusing}
Zhu, D., Zhang, T., Jiang, X., Hu, X., Chen, H., Yang, N., Lv, J., Han, J.,
  Guo, L., and Liu, T. (2014).
\newblock Fusing dti and fmri data: a survey of methods and applications.
\newblock {\em NeuroImage}, 102:184--191.

\bibitem[Zou and Hastie, 2005]{zou2005regularization}
Zou, H. and Hastie, T. (2005).
\newblock Regularization and variable selection via the elastic net.
\newblock {\em Journal of the royal statistical society: series B (statistical
  methodology)}, 67(2):301--320.

\end{thebibliography}
